\documentclass[10pt,twocolumn,letterpaper]{article}

\usepackage{wacv}
\usepackage[accsupp]{axessibility} 
\usepackage{times}
\usepackage{epsfig}
\usepackage{graphicx}
\usepackage{amsmath}
\usepackage{amssymb}
\usepackage{booktabs}
\usepackage{multirow}
\usepackage{algorithm,algorithmic}
\newcommand{\comment}[1]{}
\def\wacvPaperID{135} % Enter the WACV Paper ID here

\wacvalgorithmstrack   % Uncomment this line if you are submitting to the Algorithms Track.
\wacvfinalcopy % *** Uncomment this line for the final submission

\ifwacvfinal
\usepackage[breaklinks=true,bookmarks=false]{hyperref}
\else
\usepackage[pagebackref=true,breaklinks=true,colorlinks,bookmarks=false]{hyperref}
\fi

\def\mathbi#1{\textbf{\em #1}}
 
\begin{document}

%%%%%%%%% TITLE
\title{AVE-CLIP: AudioCLIP-based Multi-window Temporal Transformer for \underline{A}udio \underline{V}isual \underline{E}vent Localization}

\author{Tanvir Mahmud\\
The University of Texas at Austin\\
{\tt\small tanvirmahmud@uetxas.edu}
\and
Diana Marculescu\\
The University of Texas at Austin\\
{\tt\small dianam@utexas.edu}
}

\maketitle
\thispagestyle{empty}

%%%%%%%%% ABSTRACT
\begin{abstract}
An audio-visual event (AVE) is denoted by the correspondence of the visual and auditory signals in a video segment. 
Precise localization of the AVEs is very challenging since it demands effective multi-modal feature correspondence to ground the short and long range temporal interactions. 
Existing approaches struggle in capturing the different scales of multi-modal interaction due to ineffective multi-modal training strategies. 
To overcome this limitation, we introduce AVE-CLIP, a novel framework that integrates the AudioCLIP  pre-trained on large-scale audio-visual data with a multi-window temporal transformer to effectively operate on different temporal scales of video frames.
Our contributions are three-fold:
(1) We introduce a multi-stage training framework to incorporate AudioCLIP pre-trained with audio-image pairs into the AVE localization task on video frames through contrastive fine-tuning, effective mean video feature extraction, and multi-scale training phases.
(2) We propose a multi-domain attention mechanism that operates on both temporal and feature domains over varying timescales to fuse the local and global feature variations.
(3) We introduce a temporal refining scheme with event-guided attention followed by a simple-yet-effective post processing step to handle significant variations of the background over diverse events.
Our method achieves state-of-the-art performance on the publicly available AVE dataset with 5.9\% mean accuracy improvement which proves its superiority over existing approaches.
\end{abstract}

\section{Introduction}
%why do study...what is AVE

\begin{figure}[t]
  \centering
%  \fbox{\rule{0pt}{1in} \rule{0.9\linewidth}{0pt}}
   \includegraphics[width=1\linewidth]{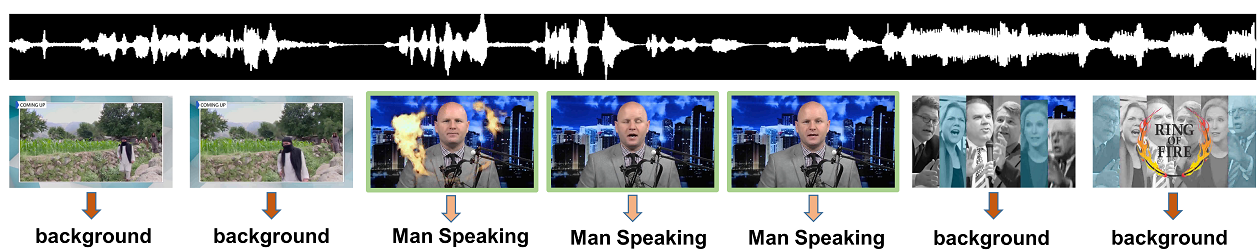}

   \caption{Example of an audio-visual event (AVE) representing the  event of an individual speaking. The person's voice is audible in all of the frames. Only when the person is visible, an AVE is identified.}
   \label{f1}
\end{figure}

Temporal reasoning of multi-modal data plays a significant role in human perception 
in diverse environmental conditions~\cite{perception, zhu}. 
Grounding the multi-modal context is critical to current and future tasks of interest, especially those that guide current research efforts in this space, \eg, embodied perception of automated agents~\cite{xia, das, gan1}, human-robot interaction with multi-sensor guidance~\cite{tsi, du, human}, and active sound source localization~\cite{zhao, arda, qian, wu2}. Similarly, audio-visual event (AVE) localization demands complex multi-modal correspondence of grounded audio-visual perception~\cite{tian, duan}. The simultaneous presence of the audio-visual cues over a video frame denotes an audio-visual event.   As shown in Fig.~\ref{f1}, the speech of the person is audible in all of the frames. However, the individual speaking  
is visible in only a few particular frames which represent the AVE.
Precise detection of such events greatly depends on the contextual understanding of the multi-modal features over the video frame.

Learning the inter-modal audio-visual feature correspondence over the video frames is one of the major challenges of AVE localization. Effective multi-modal training strategies can significantly improve performance by enhancing the relevant features. Earlier work integrates audio and image encoders pre-trained on large scale unimodal (image/audio) datasets~\cite{imagenet, audioset} to improve performance~\cite{zhou2, lin, xuan, duan, xu}. However, such a uni-modal pre-training scheme struggles to extract relevant inter-modal features that are particularly significant for AVEs. Recently, following the wide-spread success of CLIP~\cite{clip} pre-trained on large-scale vision-language datasets, AudioCLIP~\cite{audioclip} has integrated an audio encoder into the vision-language models 
with large-scale pre-training on audio-image pairs. To enhance the audio-visual feature correspondence for AVEs, we integrate the image and audio encoders from AudioCLIP with effective contrastive fine-tuning that exploits the large-scale pre-trained knowledge from multi-modal datasets instead of uni-modal ones.

Effective audio-visual fusion for multi-modal reasoning over entire video frames is another major challenge for proper utilization of the uni-modal features.
%Some earlier work~\cite{tian, lin2} focuses on  LSTM-guided sequential fusion with simple concatenation that provides sub-optimal performance because of the lack of multi-modal feature interaction. More 
Recently, several approaches have focused on 
using the grounded multi-modal features to generate temporal attention for operating
on the intra-modal feature space~\cite{zhou,lin, xuan}. Other recent work has applied recursive temporal attention on the aggregated multi-modal features~\cite{duan, lin, xu2}.
However, these existing approaches attempt to generalize audio-visual context over the whole video frame and hence struggle to extract local variational patterns that are particularly significant at event transitions.
Though generalized multi-modal context over long intervals is of great importance for categorizing diverse events, local changes of multi-modal features are critical for precise event detection at transition edges. To solve this dilemma, we introduce a multi-window temporal transformer based fusion scheme that operates on different timescales to guide attention over sharp local changes with short temporal windows, as well as extract the global context across long temporal windows.
%Such an approach greatly leverages the event classification performance while simultaneously improving detection at the transition edges.

%However, such approaches mostly guide  multi-modal fusion with temporal attentions without considering other feature domains. For better re-calibration of the aggregated feature maps, we introduce multi-domain attention over both feature and temporal that effectively strengthens the event relevant features in fusion.

%From a different perspective, existing approaches attempt to generalize context over the whole video frame and hence struggle to extract local variational patterns of the event transitions~\cite{tian, zhou, xuan2, duan}.
%Though generalized multi-modal context over long intervals is of great importance for categorizing diverse events, local changes of multi-modal features are particularly critical for precise detection at the transition edges. To solve this dilemma, we introduce a multi-window fusion scheme that operates on different timescales to guide attentions over sharp local changes with short temporal windows, as well as extract the global context across long temporal windows.Such an approach greatly leverages the event classification performance while simultaneously improving detection at the transition edges.

The background class representing uncorrelated audio-visual frames varies a lot over different AVEs for diverse surroundings (Figure~\ref{f1}).
%As shown in Figure~\ref{f1}, the background class is represented by lack of audio-visual correspondence that contains wide range of variations introduced by the surroundings.
In many cases, it becomes difficult to distinguish the background from the event regions due to subtle variations~\cite{zhou}. Xu \etal~\cite{xu} suggests that joint binary classification of the event regions (event/background) along with the multi-class event prediction improves overall performance for better discrimination of the event oriented features. Inspired by this, we introduce a temporal feature refining scheme for guiding temporal attentions over the event regions to introduce sharp contrast with the background. Moreover, we introduce a simple post-processing algorithm that filters out such incorrect predictions in between event transitions by exploiting the  high temporal locality of event/background frames in AVEs (Figure~\ref{f1}). By unifying these strategies in the AVE-CLIP framework, we achieve state-of-the-art performance on the AVE dataset which outperforms existing approaches by a considerable margin.

%considering the sequential nature of AVEs, it is expected that sequence of event/background frames is characterized by high temporal locality (Figure~\ref{f1}). This approach improves detection performance by splitting the multi-class event detection task into sequential event-background discrimination and event feature enhancement objectives. considering the sequential nature of AVEs, it is expected that sequence of event/background frames is characterized by high temporal locality (Figure~\ref{f1}).  Therefore, in most cases, the isolated AVE predictions can be treated as anomalies, and thus assumed to be incorrect predictions. To exploit this phenomenon, we introduce a simple post-processing algorithm that filters out such incorrect predictions in between event transitions. By unifying these strategies in the AVE-CLIP architecture, we achieved state-of-the-art performance on the AVE dataset that outperforms existing approaches by a considerable margin. 

The major contributions of this work are summarized as follows:

\begin{itemize}
    \item We introduce AVE-CLIP to exploit AudioCLIP pre-trained on large-scale audio-image pairs for improving inter-modal feature correspondence on video AVEs.

    \item We propose a multi-window temporal transformer based fusion scheme that operates on different timescales of AVE frames to extract local and global variations of multi-modal features.
    
%    \item We introduce a multi-domain attention mechanism to combine feature with temporal domain attention in multi-window fusion.
    
    \item We introduce a temporal feature refinement scheme through event guided temporal attention followed by a simple, yet-effective post-processing method to increase contrast with the background.
    
\end{itemize}

%multi window fusion

%two domain attention

%event guided attention framework on multi-objective learning

%simple yet effective post-processing

%state of the art performance
\begin{figure*}[t]
  \centering
  \includegraphics[width=0.8\linewidth]{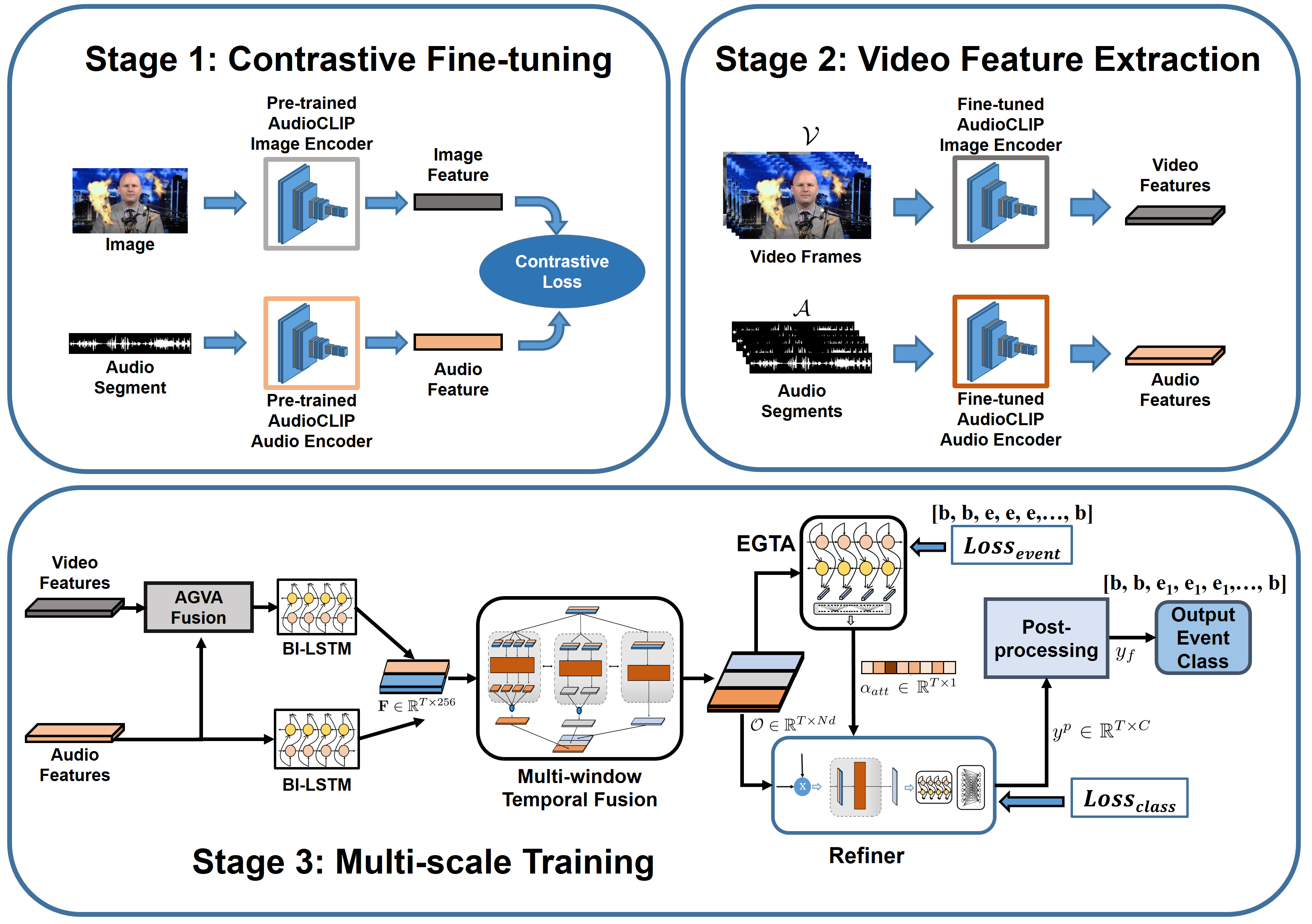}

   \caption{Schematic representation of the proposed method. In stage 1, contrastive fine-tuning is carried on the pre-trained AudioCLIP~\cite{audioclip} image and audio encoders with audio-image pairs. In stage 2, video and audio features are extracted with the fine-tuned enocoders. In stage 3, multi-scale training is carried at various temporal scales with the proposed multi-window temporal fusion module followed by temporal event refinement and post-processing to enhance event detection.}
   \label{f2}
\end{figure*}

\section{Related Work}
\subsubsection*{Audio Visual Event Localization}
AVE localization, introduced by Tian \etal~\cite{tian} targets the identification of different types of events (\eg, individual man/woman speaking, crying babies, frying food, musical instruments, \textit{etc}.) at each temporal instance based on audio-visual correspondence. The authors introduced a residual learning method with LSTM guided audio-visual attention relying on  simple concatenation and addition fusion. A dual attention matching (DAM) module is introduced by Wu \etal~\cite{wu} for operating on event-relevant features. Zhou \etal~\cite{zhou} proposed a positive sample propagation scheme by pruning out the weaker multi-modal interactions. Xuan \etal~\cite{xuan, xuan2} proposed a discriminative multi-modal attention module for sequential learning with an eigen-value based objective function. Duan \etal~\cite{duan} introduced joint co-learning with cyclic attention over the aggregated multi-modal features. Lin and Wang~\cite{lin} introduced a transformer-based approach that operates on groups of video frames based on audio-visual attention. Xu \etal~\cite{xu} introduced multi-modal relation-aware audio-visual representation learning with an interaction module. Different from existing approaches, AVE-CLIP  exploits temporal features from various windows by extracting short and long range multi-modal interactions along with temporal refinement of the event frames.

\subsubsection*{Sound Source Localization}
The sound source localization task~\cite{zhao} identifies the sounding object in the corresponding video based on the auditory signal. 
%Despite its similarity with the AVE localization, it is different since it actually depends on the correspondence of the target visual regions with the audio signal rather than detecting events. 
Arda \etal~\cite{arda} introduced an audio-visual classification model that can be adapted for sound source localization without explicit training by utilizing simple multi-modal attention. Wu \etal~\cite{wu2} proposed an encoder-decoder based framework to operate on the continuous feature space through likelihood measurements of the sounding sources. Qian \etal~\cite{qian} attempted multiple source localization by exploiting gradient weighted class activation map (Grad-CAM) correspondence on the audio-visual signal.
A self-supervised audio-visual matching scheme is introduced by Hu \etal~\cite{hu}  with a dictionary learning of the sounding objects.
Afouras \etal~\cite{afouras} utilized optical flow features along with multimodal attention maps targeting both source localization and audio source separation.

\subsubsection*{Large Scale Contrastive Pre-training}
To improve the data-efficiency on diverse target tasks, large-scale pre-training of very deep neural networks has been found to be effective for transfer learning~\cite{tl}. CLIP has introduced vision-language pre-training with self-supervised contrastive learning on large-scale datasets, an approach that received great attention for achieving superior performance on numerous multimodal vision-language tasks~\cite{clip1, clip2, clip3}. Recently, AudioCLIP~\cite{audioclip}
has extended the existing CLIP framework by integrating the audio modality with large-scale training utilizing audio-image pairs~\cite{audioset}. Such large-scale pre-training on audio-visual data can be very effective for enhancing multi-modal feature correspondence. 

%\subsubsection*{Visual Sound Separation}
%6 references
%Visual sound separation represents differentiating sounds of different sources from the input video frames. Xu \etal~\cite{xu} introduced a recursive approach for separating sound sources utilizing average energy. Gan \etal~\cite{gan} introduced graph convolutional network (GCN) to integrate visual semantic context with body dynamicsfor audio-visual separation. Zhou \etal~\cite{zhou2} introduced a multi-task learning of source separation and audio generation with a pyramid architecture for audio-visual fusion. Tian \etal~\cite{tian2} proposed a cyclic co-learning scheme by visual grounding of the audible sounding objects. Majumder \etal~\cite{majumder} introduced a reinforcement learning based approach by active exploration of the agent in the visual environment.

\section{Proposed Method}
In this paper, we introduce AVE-CLIP, a framework that integrates image and audio encoders from AudioCLIP with a multi-window temporal transformer based fusion scheme for AVE localization. Our method comprises three training stages, as presented in Figure~\ref{f2}. Initially, we start with the pre-trained weights of image and audio encoders from AudioCLIP. In stage 1, we extract image and audio segments of corresponding events to initiate fine-tuning of the pre-trained encoders on target AVE-localization frames (Section~\ref{3.2}). In stage 2, these fine-tuned encoders are deployed to extract the video and audio features from successive video frames and audio segments, respectively (Section~\ref{3.3}). Later, in stage 3, we introduce the multi-scale training on the extracted audio and video features with the multi-window temporal fusion (MWTF) module that operates on different temporal windows for generalizing the local and global temporal context (Section~\ref{3.4}).  
This is followed by the temporal refinement of the fusion feature through event-guided temporal attention generated with event-label supervision (Section~\ref{3.5}) along with a hybrid loss function used in training (Section~\ref{3.6}) and a simple post-processing algorithm that primarily enhances prediction performance during inference by exploiting the temporal locality of the AVEs (Section~\ref{3.7}).

%In Section~\ref{3.3}, we explore the early fusion modules that rely on the pre-trained feature extractors for temporal aggregation of the multi-modal features. Later, in Section~\ref{3.4}, we introduce the multi-window temporal fusion (MWTF) module that operates on different temporal windows for generalizing the local and global temporal context over the video frame. In addition, we propose a multi-domain attention scheme in the MWTF module to merge temporal and feature attention for re-calibrating the feature space (Figure~\ref{f3}). We introduce in Section~\ref{3.5} the temporal refinement of the fusion feature through event-guided temporal attention generated with event-label supervision. Afterwards, we introduce the hybrid loss function in Section~\ref{3.7} used in training. Finally, in Section~\ref{3.8}, we propose a simple post-processing algorithm that primarily enhances prediction performance during inference by exploiting the temporal locality of the AVEs.

\subsection{Preliminary}
\label{3.1}

Given a video sequence $\mathbf{S}$ of duration $T$, a set of non-overlapping video segments $\mathcal{V} = \{V^{(1)}, V^{(2)}, \dots, V^{(T)}\}$, and synchronized audio segments $\mathcal{A} = \{A^{(1)}, A^{(2)}, \dots, A^{(T)}\}$ of duration $t$ are extracted. Each video segment ${V}^{(i)} = \{ {{v_k}^{(i)}}\}_{k = 1}^{P}$ consists of $P$ image frames and the corresponding audio segment ${A}^{(i)} = \{ {{a_m}^{(i)}}\}_{m = 1}^{Q}$ consists of $Q$ samples, respectively. If the $i^{th}$ audio-video segment pair represents an event, it is labeled either as event $(e^{(i)} = 1)$ or background $(e^{(i)} = 0)$. Along with the generic event/background label, each segment of the entire video is labeled with a particular event category. Hence, the set of one-hot encoded labels\footnote{One-hot encoding changes the dimension over vector space over $\mathbb{R}$.} for the video sequence across $C$ categories is denoted by, $\mathbf{Y} = \{y^{(i)}\}_{i=1}^{T} \in \mathbb{R}^{T \times C}$. For example, let consider $\mathbf{Y} = \{b, b, b, e_2, e_2, e_2, e_3, e_3\}$ where $b$ represents background and $e_2, e_3$ denote an event of class $2$ and $3$, respectively. Here, we utilize the class labels ($y^{(i)}$) to generate event label ($e^{(i)}$) as $\{ 0, 0, 0, 1, 1, 1, 1, 1\}$ to distinguish between event $(1)$ and background $(0)$.

\begin{figure}[t]
  \centering
%  \fbox{\rule{0pt}{2in} \rule{0.9\linewidth}{0pt}}
   \includegraphics[width=0.9\linewidth]{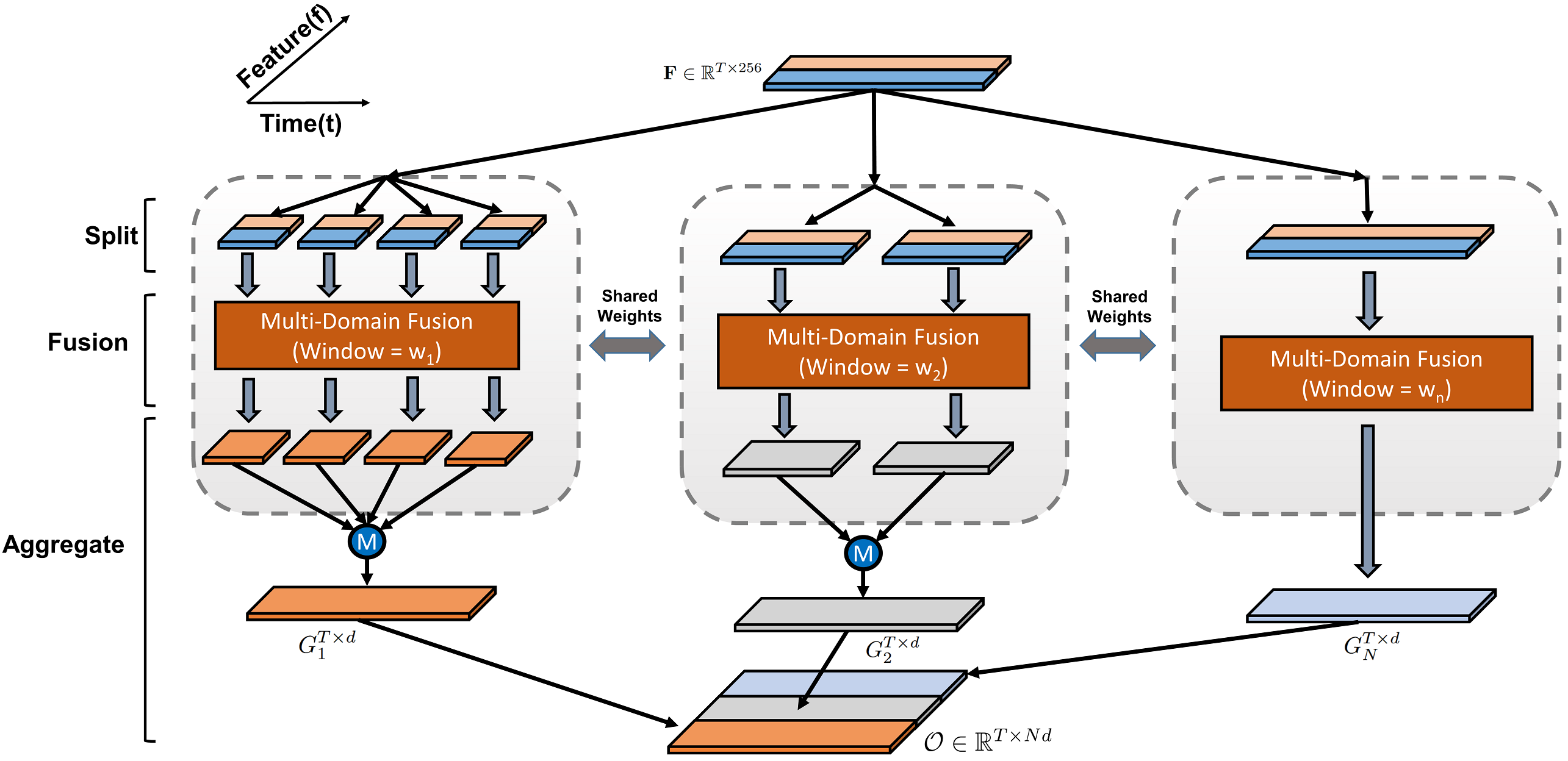}

   \caption{The three phases of the Multi-window Temporal Fusion (MWTF) module. In the \textbf{split} phase, the aggregated features are divided into separate temporal blocks based on the window length. In the \textbf{fusion} phase, multi-domain fusion is carried out on particular window. In the \textbf{aggregation} phase, temporal merging (`M') followed by feature concatenation is carried out. The window length can be varied and shared weights can be used in all fusion modules irrespective of window lengths.}
   \label{f3}
\end{figure}
\begin{figure}[t]
  \centering
%  \fbox{\rule{0pt}{2in} \rule{0.9\linewidth}{0pt}}
   \includegraphics[width=1\linewidth]{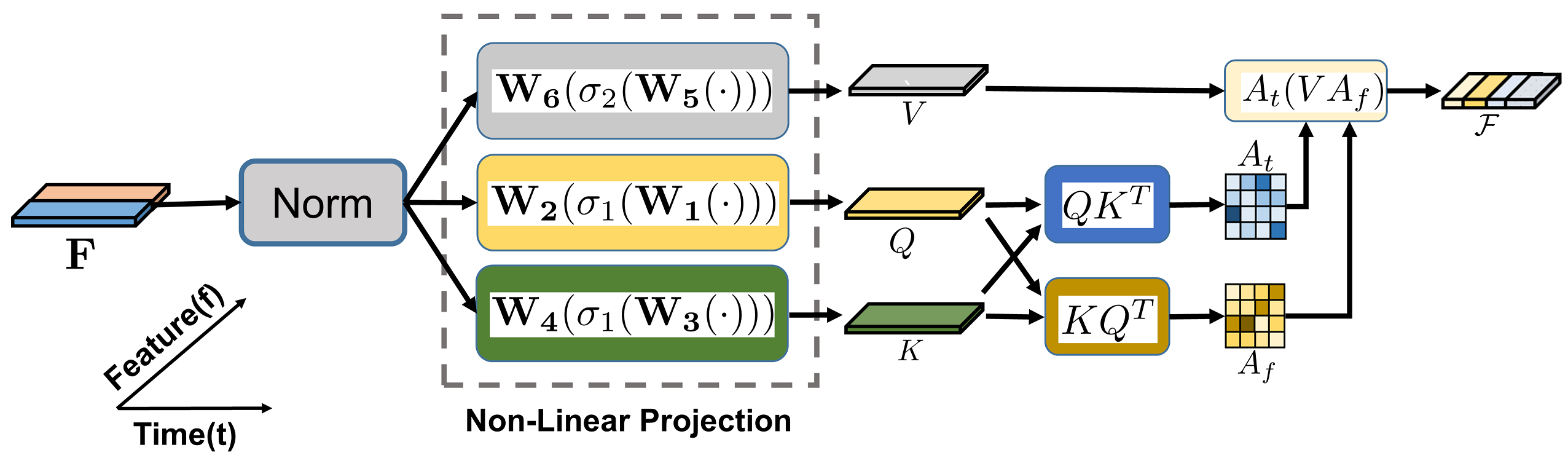}

   \caption{Representations of the\textbf{ Multi-domain Fusion} process. The attention maps are generated for both of the temporal and feature axis that are applied on the input features through non-linear projection.}
   \label{f4}
\end{figure}

%\subsection{Audio-Visual Feature Extraction with Early Fusion}
%\label{3.3}
%Initially, we aggregate the multi-modal audio-visual features by utilizing some pre-trained feature extractors followed by early fusion. A pre-trained convolutional neural network (CNN) is incorporated over each video segment $V^{(i)}$ of the entire video to generate video representation $\mathbf{s_v} \in \mathbb{R}^{T \times 7 \times 7 \times 512}$ with spatial resolution of $(7 \times 7)$ with $512$ dimensional features. Similarly, each audio segment $A^{(i)}$ is converted into spectogram representation to operate with a pre-trained CNN for generating an entire audio feature $\mathbf{s_a} \in \mathbb{R}^{T \times 128}$. To re-scale the video feature representation ($\mathbf{s_v}$) in accordance with the audio representation ($\mathbf{s_a}$) for early fusion, we adopted the Audio-Guided Video Attention (AGVA) module presented before~\cite{tian, zhou,duan}. The re-scaled video feature $\mathbf{s_{v, a}} \in \mathbb{R}^{T \times 128}$ and corresponding audio feature $\mathbf{s_a} \in \mathbb{R}^{T \times 128}$ are processed with separate bidirectional long short term memory (BiLSTM) layers for generating $\mathbf{v} \in \mathbb{R}^{T \times 128}$ and $\mathbf{a} \in \mathbb{R}^{T \times 128}$, respectively. Afterwards, temporal aggregation of audio-visual features is carried out to generate $\mathbf{F} \in \mathbb{R}^{T \times 256}$.

\subsection{Contrastive Fine-Tuning on Audio-Image Pairs}
\label{3.2}
We extract positive and negative audio-image pairs from the target dataset where a positive pair corresponds to the same AVE and a negative one represents a mismatch. Initially, we start with the pre-trained Audio and Image encoders from AudioCLIP~\cite{audioclip}. Afterwards, we intiate fine-tuning on the extracted audio-image pairs utilizing the InfoNCE loss $L_{\text{InfoNCE}} = L_I + L_A$, where $L_I$ represents the image-to-audio matching loss and $L_A$ represents the audio-to-image matching loss. $L_I$ is given by

\begin{equation}
    L_I = -\frac{1}{B} \sum_{i=1}^{B}\sum_{j=1}^{B} I_{ij} \text{log} \frac{\text{exp}(({\mathbf{z}_i^I}^\text{T} \mathbf{z}_j^A)/\tau) }{\sum_{k=1}^B \text{exp}(({\mathbf{z}_i^I}^\text{T} \mathbf{z}_j^A)/\tau)}
\end{equation}
where $B$ denotes total number of audio-image pairs in a batch, $\mathbf{z}_i^A$, $\mathbf{z}_i^I$ represent normalized audio and image features of $i^{\text{th}}$ pair, respectively, $I_{ij}$ represents the identity matrix element with $ I_{ii} = 1, \forall i = j$ and $ I_{ij} = 0, \forall i \ne j$, and $\tau$ is a trainable temperature. Similarly, we construct the audio-to-image matching loss $L_A$.

\subsection{Video Feature Extraction}
\label{3.3}
The fine-tuned audio and image encoders are deployed to extract the features from the whole video sequence $S$. To generate feature map $\mathbf{v}^{i}$ from each video segment $\mathcal{V}^{(i)}$ containing $K$ image frames, we take the mean of feature maps $\mathbf{z}_k^I \ \text{for} \ k=1, \dots, T$. Afterwards, all feature maps from $T$ video segments are concatenated to generate the video feature ${s_v}$ of a particular sequence $S$. Similarly, audio feature of each segment $\mathbf{a^{i}}$ are concatenated to generate audio feature $\mathbf{s_a}$ of a video sequence, and thus: 
\begin{align}
    \mathbf{s_v} = \mathbf{v}^{1} \oplus \mathbf{v}^{2} \oplus \dots \oplus \mathbf{v}^{T}
    \\
    \mathbf{s_a} = \mathbf{a}^{1} \oplus \mathbf{a}^{2} \oplus \dots \oplus \mathbf{a}^{T}
\end{align}

where $\oplus$ denotes feature concatenation and $T$ denotes the number of segments in a sequence.

\subsection{Multi-scale Training with Multi-window Temporal Fusion (MWTF) Transformer}
\label{3.4}
For better discrimination of the local feature variations particularly at the event transition edges, it is required to fuse multi-modal features over short temporal windows. However, the general context of the entire video is essential for better event classification.
%Therefore, it is necessary to generalize the event-context from
The proposed Multi-Window Temporal Fusion (MWTF) module effectively solves this issue by incorporating multi-domain attention over various temporal scales of the entire video, an approach that addresses the impact of both local and global variations (Figure ~\ref{f3}). 
%The operations performed in the MWTF module are schematically presented in Fig.~\ref{f3}.

Initially, to re-scale the video feature representation ($\mathbf{s_v}$) in accordance with the audio representation ($\mathbf{s_a}$) for early fusion, we adopt the Audio-Guided Video Attention (AGVA) module presented before~\cite{tian, zhou,duan}. The re-scaled video feature $\mathbf{s_{v, a}} \in \mathbb{R}^{T \times 1024}$ and corresponding audio feature $\mathbf{s_a} \in \mathbb{R}^{T \times 1024}$ are processed with separate bidirectional long short term memory (BiLSTM) layers for generating $\mathbf{v} \in \mathbb{R}^{T \times 256}$ and $\mathbf{a} \in \mathbb{R}^{T \times 256}$, respectively. Afterwards, temporal aggregation of audio-visual features is carried out to generate $\mathbf{F} \in \mathbb{R}^{T \times 512}$.

In the MWTF module, we incorporate $N$ sub-modules that operate on different timescales depending on the window length, $\{w_i \in \mathbb{R}\}_{i=1}^N$. The basic operations in each sub-module are divided into three stages: split, fusion, and aggregation. In the \textbf{split} phase of the $i^{th}$ sub-module, the aggregated feature $\mathbf{F}$ is segmented into $c_i$ blocks based on window length $w_i (w_i \times c_i = T)$ that generates $\{{F_i^{(k)} \in \mathbb{R}^{w_i \times 512}}\}_{k = 1}^{c_i}$, and thus,

\begin{equation}
    F_i^{(k)} = \mathbf{F}[ w_i(k-1) : w_ik];\ \forall k \in \{1, 2, \dots, c_i\}
\end{equation}
%Moreover, it is possible to apply varying window lengths $w$ in a sub-module, such that $\sum_{k = 1}^{c} (w_k) = T$.
In addition, it is possible to use varying window lengths in a sub-module totaling the number of time steps $T$.

Following the split action, the multi-domain attention guided \textbf{fusion} operation is carried out on each $k^{th}$ feature block $F_i^{(k)}$ of the $i^{th}$ sub-module. The multi-domain attention operation is illustrated in Figure~\ref{f4}. Considering the two-domain distribution of each block $F_i^{(k)}$, we introduce joint temporal (TA) and feature attention (FA) mechanisms by reusing the weights of similar transformations.

Firstly, each block of $F_i^{(k)} \in \mathbb{R}^{w_i \times 512}$  features is transformed to generate query vector $Q_i^{(k)} \in \mathbb{R}^{w_i \times d}$, key vector $K_i^{(k)} \in \mathbb{R}^{w_i \times d}$, and value vector $V_i^{(k)} \in \mathbb{V}^{w_i \times d}$, such that,

\begin{align}
&    Q_i^{(k)} = \mathbf{W_2}(\sigma_1 (\mathbf{W_1}(Norm(F_i^{(k)})))
\\
 &   K_i^{(k)} = \mathbf{W_4}(\sigma_1 (\mathbf{W_3}(Norm(F_i^{(k)})))
    \\
  &  V_i^{(k)} = \mathbf{W_6}(\sigma_2 (\mathbf{W_5}(Norm(F_i^{(k)})))
    \\    
    \nonumber
 & \forall i \in \{ 1, \dots, N\},\ k \in \{1, \dots, c_i\}
\end{align}

where $\mathbf{W_1} \in \mathbb{R}^{d \times 256}$, $\mathbf{W_2} \in \mathbb{R}^{d \times d}$, $\mathbf{W_3} \in \mathbb{R}^{d \times 256}$, $\mathbf{W_4} \in \mathbb{R}^{d \times d}$,  $\mathbf{W_5} \in \mathbb{R}^{d \times 256}$, $\mathbf{W_6} \in \mathbb{R}^{d \times d}$, and $\sigma_1(\cdot)$, $\sigma_2(\cdot)$  denote the $tanh$ and $Relu$ activation, respectively.

Afterwards, we process each query $Q_i^{(k)}$ and key $K_i^{(k)}$ vectors on the temporal and feature domains to generate $\beta_{t, i}^{(k)} \in \mathbb{R}^{w_i \times w_i}$ and ${\beta_{f, i}}^{(k)} \in \mathbb{R}^{d \times d}$, respectively, such that

\begin{align}
    & \beta_{t, i}^{(k)} =(QK^T)_i^{(k)} / \sqrt{d} \\
    & \beta_{f, i}^{(k)} = (KQ^T)_i^{(k)} / \sqrt{w_i} \\
    \nonumber
  & \forall i \in \{ 1, \dots, N\},\ k \in \{1, \dots, c_i\}
\end{align}

Then, we generate the temporal attention map $A_{t, i}^{(k)} \in \mathbb{R}^{w_i \times w_i}$ by applying  $softmax$ over the rows ($axis = 1$) of $\beta_{t, i}^{(k)} \in \mathbb{R}^{w_i \times w_i}$. In addition, the feature attention map $A_{f, i}^{(k)} \in \mathbb{R}^{d \times d}$ is generated by applying $softmax$ over the columns ($axis= 0$) of $\beta_{f, i}^{(k)} \in \mathbb{R}^{d \times d}$.

These multi-domain attention maps are sequentially applied on each axis of $V_i^{(k)}$ to generate the modified feature map $\mathcal{F}_i^{(k)} \in \mathbb{R}^{w_i \times d}$ by,
\begin{align}
    & \mathcal{F}_i^{(k)} = A_{t, i}^{(k)} (V_{i}^{(k)} A_{f, i}^{(k)})\\
    & \nonumber \forall i \in \{ 1, \dots, N\},\ k \in \{1, \dots, c_i\}
\end{align}

Finally, in the \textbf{aggregate} phase, the modified feature maps of each $i^{th}$ block in a sub-module are temporally concatenated to generate $G_i \in \mathbb{R}^{T \times d}$ such that,

\begin{equation}
G_i = (\mathcal{F}_i^{(1)} || \mathcal{F}_i^{(2)} || \dots || \mathcal{F}_i^{(c_i)}); \    \forall i \in \{ 1, \dots, N\}
\end{equation}
where `$||$' denotes the feature concatenation along temporal axis.

Afterwards, all the modified feature maps from each sub-module are concatenated along channel axis maintaining temporal relation  to generate $\mathcal{O}\in \mathbb{R}^{T \times Nd}$ as,

\begin{equation}
    \mathcal{O} =  G_1 \oplus G_2 \oplus \dots \oplus G_{c_i}
\end{equation}
where $\oplus$ denotes feature concatenation along channel axis.

\subsection{Event-Guided Temporal Refinement}
\label{3.5}
As the background class represents miss-aligned audio-visual pairs from all other classes, it often becomes difficult to distinguish them from event classes in case of subtle variations. To enhance the contrast between the event-segments and backgrounds, we introduce supervised event-guided  temporal attention (EGTA) over the event region. Following EGTA, we refine the contrasted event segments with an additional stage of single window fusion for better discrimination of the event categories.

To generate EGTA $\alpha_{att} \in \mathbb{R}^{T \times 1}$,  the fusion vector $\mathcal{O}$ is passed through a BiLSTM module as,
\begin{align}
    & \gamma = BiLSTM(Norm(\mathcal{O}))
    \\
    & {\alpha_{att}} = \sigma_3 (\mathbf{W_7}(\gamma))
\end{align}
where $\gamma \in \mathbb{R}^{T \times d'}$, $\mathbf{W_7} \in \mathbb{R}^{1 \times d'}$, and $\sigma_3(\cdot)$ represents the $sigmoid$ activation function. 

Afterwards, in the refining phase, we apply the EGTA mask ${\alpha_{att}}$ over the fusion vector $\mathcal{O}$ to generate $\mathcal{O'} \in \mathbb{R}^{T \times Nd}$ by,
\begin{equation}
    \mathcal{O}' = \mathcal{O} \odot (\alpha_{\text{att}} \mathbi{1}) 
\end{equation}
where $\mathbi{1} \in \mathbb{R}^{1 \times Nd}$ denotes a broadcasting vector with all ones, and $\odot$ represents the element-wise multiplication.

As a last step, we incorporate a single window fusion with window length $w = T$ to generate $\mathcal{O}_w \in \mathbb{R}^{T \times d}$ by refining event-concentrated vector $\mathcal{O'}$. Finally, we obtain the final event category prediction ${y}^p \in \mathbb{R}^{T \times C}$ after applying another sequential BiLSTM layer as,

\begin{equation}
    {y^p} = softmax(\mathbf{W_8}(BiLSTM(Norm(\mathcal{O}_w))))
\end{equation}
where $\mathbf{W_8} \in \mathbb{R}^{C \times d}$ for $C$ categories of AVE.

%%%%%%%%%%
\subsection{Loss Function}
\label{3.6}
To guide the event-attention $(\alpha_{att})$ for refinement, we use an event label loss $\mathcal{L}_e$. Also, for the multi-class prediction $y^p$, an event category loss $\mathcal{L}_c$ is incorporated. The total loss ${L}$ is obtained by combining $\mathcal{L}_e$ and $\mathcal{L}_c$ as,

\begin{align}
    {L} =  \lambda_1 \overbrace{\mathcal{L}_{BCE} {(\alpha_{att}, E)}}^{\mathcal{L}_e} + \lambda_2 \overbrace{\mathcal{L}_{CE} {(y^p, y)}}^{\mathcal{L}_c}
\end{align}
where $\lambda_1$, $\lambda_2$ denote weighting factors, ${E} \in \mathbb{R} ^{T \times 1}$ denotes the binary event label, and ${y} \in \mathbb{R} ^{T \times C}$ denotes the one-hot encoded multi-class event categories over the time-frame.

\subsection{Post-Processing Algorithm during Inference}
\label{3.7}
Due to the sequential nature of AVEs, it is expected that they have high locality over the entire video frame. 
Therefore, AVEs are typically clustered together and isolated non-AVEs can be viewed as anomalies. 
We exploit this property to filter the generated event prediction ${y^p}$ during inference for obtaining the final prediction $y_{f}$. Here, we consider a window length $W$ to represent the minimum number of consecutive predictions required for considering any change as an anomaly. As such, all non-matching $y^p$ values are corrected according to the prevailing one.

\section{Experiments and Analysis}

\subsection{Experimental Setup}

\subsubsection*{Audio-Visual Event Dataset}
The Audio-Visual Event Dataset, introduced by Tian \etal~\cite{tian}, is widely used for the audio-visual event localization task. The dataset contains $4,143$ video clips along with the audio containing $28$ different events including daily human activities, instrument performances, animal actions, and vehicle activities. Each video clip is $10$ seconds long with temporal start/end annotations for all events. According to existing work~\cite{tian, zhou, lin}, training/validation/test splits of $3,309/ 402/ 402$ are considered for evaluation of the proposed method. 

\subsubsection*{Evaluation Metrics}
Following existing work~\cite{xuan, zhou, lin, duan}, we consider the final classification accuracy of multi-class events over the entire video as the evaluation metric. Along with the background, $29$ event classes are considered for per-second prediction over the $10s$ video duration where the video sampling rate varies from $16$ to $48$. The background category includes all the misaligned audio-visual segments that don't belong to any of the $28$ main categories.

\subsubsection*{Implementation Details}
We incorporate the pre-trained audio and image encoders from the AudioCLIP~\cite{audioclip} framework that fine-tunes the pre-trained CLIP~\cite{clip} framework with audio-image pairs extracted from the large-scale AudioSet~\cite{audioset} dataset. The audio encoder is the ESResNeXt model~\cite{esr} that
is based on the ResNeXt-50~\cite{resnext} architecture and the image encoder is a ResNet-50~\cite{resnet} model. We used combination of four MWTF modules for experiments that is defined empirically.
%As in existing work~\cite{xuan, lin, zhou}, we adopted the similar VGG-19 backbone pre-trained on ImageNet~\cite{imagenet} to initially extract the video features, and another VGG-like network pre-trained on AudioSet~\cite{audioset} to extract audio features from \textit{log-mel} spectogram representation. 
The weights of the hybrid loss function are empirically chosen to be $(\lambda_1 = 0.3, \lambda_2 = 0.7)$. For evaluation, each network combination is trained for $300$ epochs on $256$ AMD EPYC ${7742}$ CPUs with $2$ Quadro GV100 and $2$ A100-PCIE-40GB GPUs.

\subsection{Comparison of State-of-the-Art methods}
AVE-CLIP is compared against several state-of-the art methods  in Table~\ref{t1}. The multi-modal approaches  outperform the uni-modal ones by a great margin. This is expected given the richer context provided by the multi-modal analysis over the uni-modal counterpart.
The multi-modal audio-visual fusion strategies play a very critical role in the AVE localization performance. Traditionally, various audio-visual co-attention fusion schemes have been explored to enhance the temporal event features that provide comparable performances. Recently, Lin \etal~\cite{lin} introduced a transformer based multi-modal fusion approach that incorporates instance-level attention to follow visual context over consecutive frames. However, the proposed AVE-CLIP architecture achieves the best performance \textbf{with an accuracy of \textbf{83.7\%}} that outperforms the corresponding transformer based approach by \textbf{6.9\%}. Moreover, the AVE-CLIP provides \textbf{5.9\%} higher accuracy compared to the best-performing co-attention approach proposed by Zhou \etal~\cite{zhou}. 
%The multi-window temporal fusion approach of the AVE-CLIP greatly provides better generalization of the local and global context that results in significant performance improvement over its counterparts.

%%%%%%%%%Table 1
% \usepackage{multirow}
\begin{table}[t]
\scalebox{0.9}{
\begin{tabular}{ccc}
\toprule
\multicolumn{2}{c}{\textbf{Method}}                                                                                                                                      & \multicolumn{1}{c}{\textbf{Accuracy(\%)}} \\ \midrule
\multirow{2}{*}{uni-modal}                                                                          & Audio-based~\cite{audio}                                                   & 59.5                                       \\
                                                                                                   & Video-based~\cite{very}                                                   & 55.3                                       \\ \midrule
\multirow{5}{*}{\begin{tabular}[c]{@{}c@{}}multi-modal\\ (with Co-Attention\\ Fusion)\end{tabular}} & AVEL~\cite{tian}                                                           & 74.7                                       \\
                                                                                                   & DAM~\cite{wu}                                                            & 74.5                                       \\
                                                                                                   & PSP~\cite{zhou}                                                            & 77.8                                       \\
                                                                                                   & AVIN~\cite{ramas}                                                            & 75.2                                       \\
                                                                                                   & RFJC~\cite{duan}                                                           & 76.2                                       \\ \midrule
\multirow{2}{*}{\begin{tabular}[c]{@{}c@{}}multi-modal\\ (with Transformer\\ Fusion)\end{tabular}}  & AV-Transformer~\cite{lin}                                                           & 76.8                                       \\
                                                                                                   & \textbf{\begin{tabular}[c]{@{}c@{}}AVE-CLIP \\ (ours)\end{tabular}} & \textbf{83.7}       \\ \bottomrule

\end{tabular}}
\caption{Performance comparison of the state-of-the-art methods on AVE classification. Various uni-modal methods and multi-modal fusion strategies are compared.}
\label{t1}
\end{table}

\begin{table}[t]
\scalebox{0.9}{
\begin{tabular}{cc}
\toprule
\textbf{Image/Audio Encoders}                               & \textbf{Accuracy(\%)} \\
\midrule
with AudioClip-Encoders (w/o Fine-tuning)  & 81.1         \\
with AudioClip-Encoders (with Fine-tuning) & \textbf{83.7}         \\
without AudioCLIP-Encoders                & 79.3    \\  
\bottomrule
\end{tabular}}
\caption{Impact of the pre-trained AudioCLIP image and audio encoders with contrastive fine-tuning on the AVE-CLIP framework.}
\label{nt}
\end{table}

\subsection{Ablation Study}
To analyze the effect of individual modules in proposed AVE-CLIP, we carried out a detailed ablation study over the baseline approach. The final AVE-CLIP architecture integrates the best performing structures of the building blocks. 

%%%%%%%%%%%%%%%%%%Table 2
% \usepackage{multirow}
\begin{table}[t]
\centering
\resizebox{\columnwidth}{!}{
\begin{tabular}{clc}
\toprule
\multicolumn{2}{c}{\textbf{Strategies}}                                                                                                                                              & \textbf{Accuracy(\%)} \\ \midrule
\multirow{3}{*}{\begin{tabular}[c]{@{}c@{}}multi-modal\\ Fusion\end{tabular}} & with PSP~\cite{zhou}                                                                                      & 77.8                  \\
                                                                             & with AV-Transformer~\cite{lin}                                                                            & 76.8                  \\
                                                                             & with only MWTF and  AudioCLIP encoders (ours)                                                                                      & 82.0                  \\ \midrule
\multirow{2}{*}{\begin{tabular}[c]{@{}c@{}}Temporal\\ Refining\end{tabular}} & with MWTF + Refiner (ours)                                                                            & 82.5                  \\
                                                                             & with MWTF + EGTA + Refiner (ours)                                                                     & 83.2                  \\ \midrule
\textbf{\begin{tabular}[c]{@{}c@{}}Complete\\ Model\end{tabular}}            & \textbf{\begin{tabular}[c]{@{}l@{}}with MWTF + EGTA + Refiner\\ + Post Processing (ours)\end{tabular}} & \textbf{83.7}    \\ \bottomrule    
\end{tabular}}
\caption{Impact of various building blocks on the performance of the proposed AVE-CLIP architecture.}
\label{t2}
\end{table}
%%%%%%%%%%%%%%%%%%%%%Table 3

% \usepackage{multirow}
% \usepackage{multirow}
\begin{table}[t]
\centering
\scalebox{0.8}{
\begin{tabular}{ccccc}
\toprule
\multirow{2}{*}{\textbf{\begin{tabular}[c]{@{}c@{}}Attention\\ Axis\end{tabular}}} & \multicolumn{4}{c}{\textbf{Window Length, w}}                \\ \cmidrule{2-5}
                                                                                   & \textbf{2s} & \textbf{5s} & \textbf{10s} & \textbf{Variable} \\
                                                                                   \midrule
\textbf{Temporal}                                                                  & 78.1        & 79.3        & 81.2         & 79.4              \\
\textbf{Feature}                                                                   & 78.5        & 79.1        & 81.6         & 79.2              \\
\textbf{Multi-domain}                                                                       & 79.0        & 79.8        & \textbf{81.6}         & 79.7    \\ \bottomrule         
\end{tabular}}
\caption{Accuracy (\%) obtained with a \textbf{single} fusion block of MWTF module for different attention domains with various window lengths. The variable window length denotes a combination of $(3s, 3s, 4s)$ windows.}
\label{t3}
\end{table}

%%%%%%%%%%%%%%%%%Table 4
% \usepackage{multirow}
\begin{table}[t]
\centering
\scalebox{0.8}{
\begin{tabular}{lcc}
\toprule
\multirow{2}{*}{\textbf{\begin{tabular}[c]{@{}c@{}}Window\\ Combination\\ (with Two Att.)\end{tabular}}} & \multicolumn{2}{c}{\textbf{Accuracy(\%)}}                                                                                                  \\ \cmidrule{2-3} 
                                                                                                         & \textbf{\begin{tabular}[c]{@{}c@{}}Shared\\ Weights\end{tabular}} & \textbf{\begin{tabular}[c]{@{}c@{}}Independent\\ Weights\end{tabular}} \\ \midrule
10s                                                                                                      & -                                                                 & 81.6                                                                   \\
10s + 5s                                                                                                 & 82.4                                                              & 82.7                                                                   \\
10s + 3s*                                                                                                & 82.1                                                              & 82.5                                                                   \\
5s + 2s                                                                                                  & 80.6                                                              & 81.2                                                                   \\
10s + 5s + 3s*                                                                                           & 83.2                                                              & 82.8                                                                   \\
10s + 5s + 2s                                                                                            & 82.9                                                              & 83.0                                                                   \\
10s + 5s + 3s* + 2s                                                                                      & \textbf{83.7}                                                     & 83.3                                                                 \\ \bottomrule 
\end{tabular}}
\caption{Impact of using different combinations of window for the Multi-window Temporal Fusion (MWTF) module. `*' indicates the use of variable window lengths for $3s$.}
\label{t4}
\end{table}

\subsubsection*{Effect of Contrastive Fine-tuning in Encoders}
We incorporate pre-trained image and audio encoders into the AVE-CLIP framework from AudioCLIP that are subjected to contrastive fine-tuning in training stage $1$ (Section~\ref{3.2}). The effect of these encoders on the final performance of AVE-CLIP is summarized in Table~\ref{nt}. For the baseline comparisons,
we adopted the similar VGG-19 backbone pretrained on ImageNet~\cite{imagenet} to initially extract the video features, and another VGG-like network pre-trained on AudioSet~\cite{audioset} to extract audio features following existing work~\cite{xuan, lin, zhou}.
We observe that the best performance of $83.7\%$ is achieved by the AudioCLIP encoders with contrastive fine-tuning which improves accuracy by $4.4\%$ over the uni-modal encoders. Moreover, the contrastive fine-tuning phase improves the accuracy of AVE-CLIP by $2.6\%$ which shows its effectiveness on AVE localization. 

\subsubsection*{Effect of Multi-window Temporal Fusion (MWTF)}
For analyzing the effect of the MWTF module in AVE-CLIP, the rest of the modules (temporal refining, post-processing) are replaced with simple fully connected layers followed by a \textit{softmax} classifier. 
Moreover, to compare with other multi-modal fusion schemes, the PSP based fusion~\cite{zhou} and the AV-Transformer based fusion from~\cite{lin} are considered. 
The resulting accuracy is summarized in Table~\ref{t2}. The proposed MWTF module with AudioCLIP encoders provides significant improvements over existing counterparts which shows its effectiveness.

The MWTF module provides a generic framework to operate on various temporal resolutions with shared weights for effectively guiding the multi-modal attention. 
To analyze the effect of various temporal window lengths, the performance when using a single fusion block in MWTF module is explored in Table~\ref{t3}. The model performs better with increasing window length while achieving the best performance with $10s$ window length.
Moreover, we can observe the consistent performance improvements with multi-domain attention over their single domain counterparts.
Although the fusion scheme with a smaller attention window achieves more discriminating features emphasizing high frequency local variations, it misses the global context which is particularly significant for differentiating event categories. 
%The accuracy improvements with increasing window length are greatly influenced by the broader observation perspective that generalizes well across events.

The performance for combinations of varying window lengths is provided in Table~\ref{t4}. By incorporating these fusion windows into the $10s$ window, performance increases considerably when compared with the baseline. Despite the lower performance for smaller window lengths, these configurations are better at extracting discriminatory features which are particularly critical to determining the sharp edges of event transitions. 
Hence, the combination of larger and smaller window features is effective for generalizing global low frequency features as well as local high frequency variations at the transition edges. 
Moreover, we observe that independents weights over different fusion module perform better compared to their shared counter parts with fewer windows.
However, when the number of windows increases, such advantages appear to shrink due to the increased complexity of the fusion mechanism.

\subsubsection*{Effect of Event-Guided Temporal Feature Refining}
Downstream from the fusion module, AVE-CLIP includes temporal feature refining which consists of two phases: the event-guided temporal attention (EGTA) mask generation, and the corresponding feature refinement. From Table~\ref{t2}, it can be seen the effect of temporal refinement with EGTA which produces an improvement of $0.7\%$ in accuracy. Moreover, the performance of different combinations for feature refining is provided in Table~\ref{t6}. It is possible to generate the EGTA module without event guided supervision which, as a result, simplifies the loss function to the simple cross-entropy loss. However, with event-label supervision, the model distinguishes the event frames better against backgrounds which in turn provides better performance.
For the refiner, the single-window based fusion with $w = 10s$ generates the best performance since multi-window fusion becomes gradually saturated in this phase. 

%Due to the variations in background features, such event-label attention greatly enhances performance. The temporal attention guided features are followed by another stage of temporal refinement for improving event categorization performance. For the refiner, the single-window based fusion with $w = 10s$ generates the best performance since multi-window fusion gets gradually saturated in this phase. 

%\subsubsection*{Effect of Feature Refinement with Temporal Attention}

\subsubsection*{Effect of Post-Processing Algorithm}
Considering the sequential nature of the events, the proposed post-processing method is found to be very effective for achieving better prediction during inference. As the prediction of event category is generated on a per-second basis, incorrect predictions can be reduced by considering a window of consecutive predictions. The effects of different window lengths on the post-processing method are summarized in Table~\ref{t7}. %As it can be seen,
The best performance is achieved for a $3s$ window length. With a smaller window length, the effect of filtering is reduced over longer events whereas the larger windows reduces the performance in shorter events.

%%%%%%%%%Table 6
% \usepackage{multirow}
\begin{table}[t]
\centering
\scalebox{0.8}{
\begin{tabular}{llc}
\toprule
\multicolumn{2}{c}{\textbf{Method}}                     & \textbf{Accuracy(\%)}  \\ \midrule
\multirow{2}{*}{EGTA}    & with supervision    & \textbf{83.7}          \\
                         & without supervision & 83.1          \\ \midrule
\multirow{3}{*}{Refiner} & window = 10s        & \textbf{83.7} \\
                         & window = 5s         & 83.2          \\
                         & window = (10s +5s)  & 83.6        \\ \bottomrule 
\end{tabular}}
\caption{The effect of event label supervision on the Event-Guided Temporal Attention (EGTA) module and the effect of various fusion window lengths on the Refiner module.}
\label{t6}
\end{table}

%%%Table 5
\begin{table}[t]
\centering
\scalebox{0.8}{
\begin{tabular}{cccccc}
\toprule
\textbf{window length} & 1s   & 2s   & 3s            & 4s   & 5s   \\ \midrule
\textbf{Accuracy(\%)} & 83.2 & 83.5 & \textbf{83.7} & 82.9 & 82.3 \\ \bottomrule
\end{tabular}}
\caption{The effect of different window lengths of the post-processing module during Inference. The best performing modules are considered in AVE-CLIP architecture for the baseline.}
\label{t7}
\end{table}

%Considering the sequential nature of the events, the proposed post-processing method is found to be very effective for achieving better prediction during inference. Performances of the post-processing method with different window lengths are summarized in Table~\ref{t5}. The best performance is achieved with $3s$ window length in post-processing. Smaller window length reduces the filtering effect over longer events while larger windows gradually reduces the performance in shorter events that demands optimum window selection. 

%Performances of the post-processing method with different window lengths are summarized in Table~\ref{t5}. The best performance is achieved with $3s$ window length in post-processing. 
%Smaller window length reduces the filtering effect over longer events while larger windows gradually reduces the performance in smaller events that demands optimum window selection. 

%As the prediction of event category is generated on per-second basis, false predictions can be reduced by considering a window of subsequent predictions before altering the stream of event prediction.  

\subsection{Qualitative Analysis}
%\subsubsection*{Multi-Windowed Attention Improves Localization}

%\subsubsection*{Better Performance at the Event Transition}

%\subsubsection*{Improved Classification at Shorter Events}

%\subsubsection*{MAttNet improves feature discrimination}
   
\begin{figure}[t]
  \centering
%  \fbox{\rule{0pt}{2in} \rule{0.9\linewidth}{0pt}}
   \includegraphics[width=0.95\linewidth]{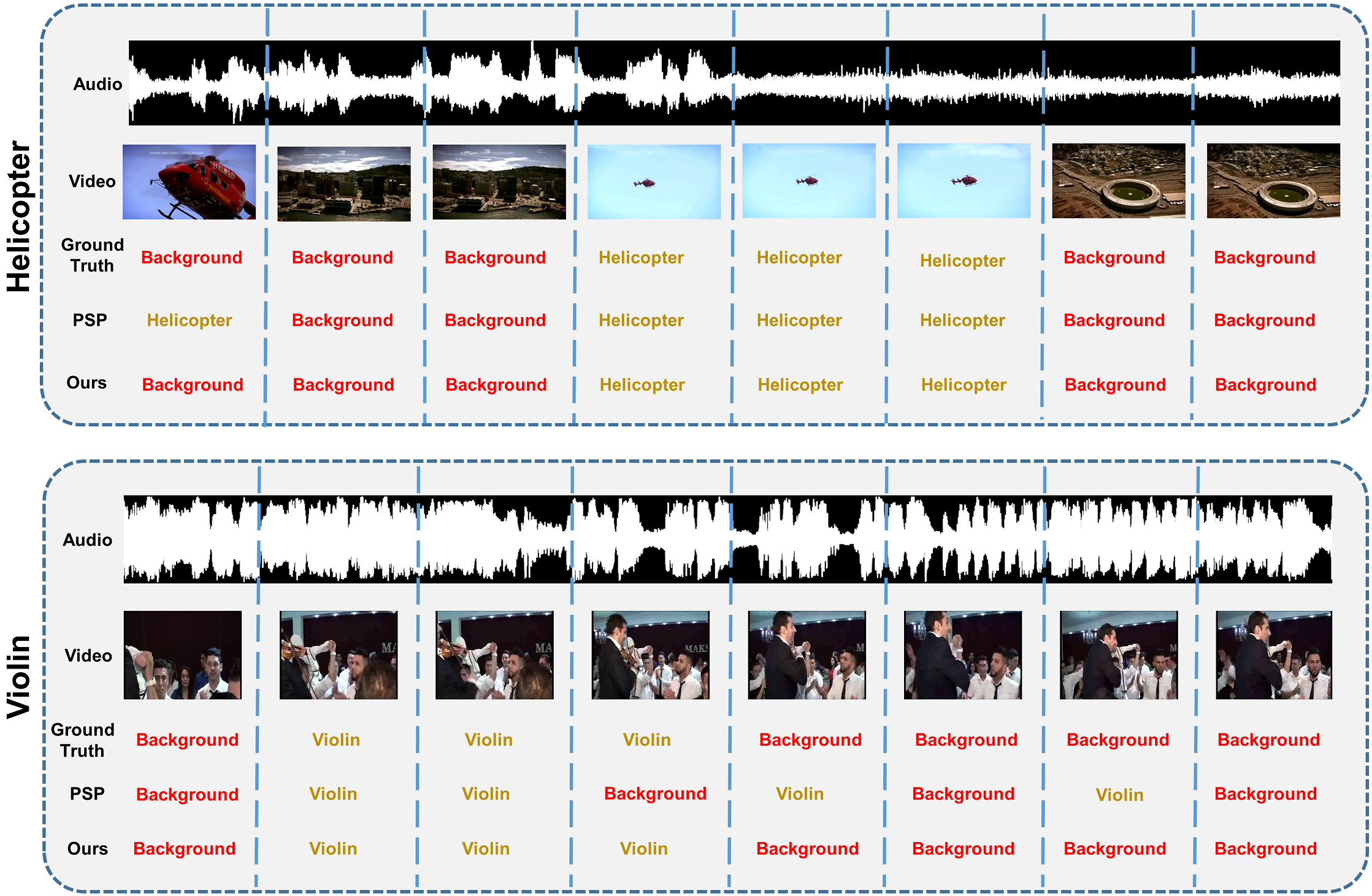}

   \caption{Visual representation of performances of PSP~(\cite{zhou}) and AVE-CLIP on two AVE events (helicopter and  violin). The AVE-CLIP performs better to localize event transitions.}
   \label{c2}
\end{figure}

%\subsubsection*{The Post Processing Reduces False Negatives at Long Events}
The qualitative performance of the proposed AVE-CLIP is demonstrated in Figure~\ref{c2} for two audio-visual events. For comparative analysis, we have shown the performance of the PSP model (\cite{zhou}) as well. In the first event, the AVE represents a moving helicopter. Though the helicopter is visible in the first frame, it is a background event due to the absence of the flying helicopter sound. Only the middle three frames capture the AVE through audio-visual correspondence. Our proposed method perfectly distinguishes the helicopter event whereas PSP (\cite{zhou}) fails at the challenging first frame. The second event representing a person playing the violin is very challenging given that the violin is hardly visible. Though the sound of violin is present throughout, the image of violin is visible in only few frames that represents the AVE. The PSP (\cite{zhou}) method generates some incorrect predictions at event transitions. However, the proposed AVE-CLIP perfectly distinguishes the event frames, which demonstrates its effectiveness for generalizing local variations. Furthermore, AVE-CLIP achieves better performance in many challenging cases that demands different scales of temporal reasoning throughout video.

%\section{Future Works}
%The proposed AVE-CLIP achieves state-of-the-art performance on supervised audio-visual event localization. The method can be easily extended to weakly-supervised localization without temporal annotation of each segment. The multi-window fusion scheme can be very effective for various audio-visual applications, \eg multi-modal correspondence, sound source localization, and video highlight detection. Apart from multi-modal applications, the concept of operating on various temporal scales should provide richer context on any uni-modal temporal learning objective either relying on audio or video modality that can be a promising direction for further research.

\section{Conclusion}
In this paper, we introduced AVE-CLIP that uses AudioCLIP encoders in conjunction with a multi-scale temporal fusion based transformer architecture for improving AVE localization performance. 
We show that the effect of AudioCLIP encoders with contrastive fine-tuning is significant in AVE-localization 
for generating improved multi-modal representation.
Our results show that local feature variations are  essential for event transition detection while global variations are critical for identifying different event classes. The proposed multi-window fusion module exploits both local and global variations with multi-domain attention thereby significantly improving performance. %We also show that the multi-domain attention mechanism operating over temporal and feature domains to be very effective in multi-window fusion. 
The temporal refining of the event frames simplifies the event classification task which improves multi-class AVE localization performance. Finally, by exploiting the sequential nature of AVEs with a simple post-processing scheme, we were able to achieve state-of-the-art performance on the AVE dataset.

\section*{Acknowledgements}
This research was supported in part by the Office of Naval Research, Minerva Program, and a UT Cockrell School of Engineering Doctoral Fellowship.

%%%%%%%%% REFERENCES
{\small
\bibliographystyle{ieee_fullname}
\bibliography{egbib}

\begin{thebibliography}{10}\itemsep=-1pt

\bibitem{afouras}
Triantafyllos Afouras, Andrew Owens, Joon~Son Chung, and Andrew Zisserman.
\newblock Self-supervised learning of audio-visual objects from video.
\newblock In {\em ECCV}, pages 208--224. Springer, 2020.

\bibitem{human}
Partha Chakraborty, Sabbir Ahmed, Mohammad~Abu Yousuf, Akm Azad, Salem~A
  Alyami, and Mohammad~Ali Moni.
\newblock A human-robot interaction system calculating visual focus of
  human’s attention level.
\newblock {\em IEEE Access}, 9:93409--93421, 2021.

\bibitem{resnext}
Fran{\c{c}}ois Chollet.
\newblock Xception: Deep learning with depthwise separable convolutions.
\newblock In {\em Proceedings of the IEEE conference on computer vision and
  pattern recognition}, pages 1251--1258, 2017.

\bibitem{das}
Abhishek Das, Samyak Datta, Georgia Gkioxari, Stefan Lee, Devi Parikh, and
  Dhruv Batra.
\newblock Embodied question answering.
\newblock In {\em CVPR}, pages 1--10, 2018.

\bibitem{imagenet}
Jia Deng, Wei Dong, Richard Socher, Li-Jia Li, Kai Li, and Li Fei-Fei.
\newblock Imagenet: A large-scale hierarchical image database.
\newblock In {\em CVPR}, pages 248--255. IEEE, 2009.

\bibitem{du}
Guanglong Du, Mingxuan Chen, Caibing Liu, Bo Zhang, and Ping Zhang.
\newblock Online robot teaching with natural human--robot interaction.
\newblock {\em IEEE Transactions on Industrial Electronics}, 65(12):9571--9581,
  2018.

\bibitem{duan}
Bin Duan, Hao Tang, Wei Wang, Ziliang Zong, Guowei Yang, and Yan Yan.
\newblock Audio-visual event localization via recursive fusion by joint
  co-attention.
\newblock In {\em WACV}, pages 4013--4022, 2021.

\bibitem{gan1}
Chuang Gan, Yiwei Zhang, Jiajun Wu, Boqing Gong, and Joshua~B Tenenbaum.
\newblock Look, listen, and act: Towards audio-visual embodied navigation.
\newblock In {\em ICRA}, pages 9701--9707. IEEE, 2020.

\bibitem{audioset}
Jort~F Gemmeke, Daniel~PW Ellis, Dylan Freedman, Aren Jansen, Wade Lawrence,
  R~Channing Moore, Manoj Plakal, and Marvin Ritter.
\newblock Audio set: An ontology and human-labeled dataset for audio events.
\newblock In {\em ICASSP}, pages 776--780. IEEE, 2017.

\bibitem{perception}
Esam Ghaleb, Mirela Popa, and Stylianos Asteriadis.
\newblock Multimodal and temporal perception of audio-visual cues for emotion
  recognition.
\newblock In {\em ACII}, pages 552--558. IEEE, 2019.

\bibitem{esr}
Andrey Guzhov, Federico Raue, J{\"o}rn Hees, and Andreas Dengel.
\newblock Esresne (x) t-fbsp: Learning robust time-frequency transformation of
  audio.
\newblock In {\em 2021 International Joint Conference on Neural Networks
  (IJCNN)}, pages 1--8. IEEE, 2021.

\bibitem{audioclip}
Andrey Guzhov, Federico Raue, J{\"o}rn Hees, and Andreas Dengel.
\newblock Audioclip: Extending clip to image, text and audio.
\newblock In {\em ICASSP 2022-2022 IEEE International Conference on Acoustics,
  Speech and Signal Processing (ICASSP)}, pages 976--980. IEEE, 2022.

\bibitem{resnet}
Kaiming He, Xiangyu Zhang, Shaoqing Ren, and Jian Sun.
\newblock Deep residual learning for image recognition.
\newblock In {\em Proceedings of the IEEE conference on computer vision and
  pattern recognition}, pages 770--778, 2016.

\bibitem{audio}
Shawn Hershey, Sourish Chaudhuri, Daniel~PW Ellis, Jort~F Gemmeke, Aren Jansen,
  R~Channing Moore, Manoj Plakal, Devin Platt, Rif~A Saurous, Bryan Seybold,
  et~al.
\newblock Cnn architectures for large-scale audio classification.
\newblock In {\em ICASSP}, pages 131--135. IEEE, 2017.

\bibitem{hu}
Di Hu, Rui Qian, Minyue Jiang, Xiao Tan, Shilei Wen, Errui Ding, Weiyao Lin,
  and Dejing Dou.
\newblock Discriminative sounding objects localization via self-supervised
  audiovisual matching.
\newblock {\em NeurIPS}, 33:10077--10087, 2020.

\bibitem{tl}
Yanghao Li, Saining Xie, Xinlei Chen, Piotr Dollar, Kaiming He, and Ross
  Girshick.
\newblock Benchmarking detection transfer learning with vision transformers.
\newblock {\em arXiv preprint arXiv:2111.11429}, 2021.

\bibitem{lin}
Yan-Bo Lin and Yu-Chiang~Frank Wang.
\newblock Audiovisual transformer with instance attention for audio-visual
  event localization.
\newblock In {\em ACCV}, 2020.

\bibitem{qian}
Rui Qian, Di Hu, Heinrich Dinkel, Mengyue Wu, Ning Xu, and Weiyao Lin.
\newblock Multiple sound sources localization from coarse to fine.
\newblock In {\em ECCV}, pages 292--308. Springer, 2020.

\bibitem{clip}
Alec Radford, Jong~Wook Kim, Chris Hallacy, Aditya Ramesh, Gabriel Goh,
  Sandhini Agarwal, Girish Sastry, Amanda Askell, Pamela Mishkin, Jack Clark,
  et~al.
\newblock Learning transferable visual models from natural language
  supervision.
\newblock In {\em International Conference on Machine Learning}, pages
  8748--8763. PMLR, 2021.

\bibitem{ramas}
Janani Ramaswamy.
\newblock What makes the sound?: A dual-modality interacting network for
  audio-visual event localization.
\newblock In {\em ICASSP}, pages 4372--4376. IEEE, 2020.

\bibitem{clip1}
Aditya Sanghi, Hang Chu, Joseph~G Lambourne, Ye Wang, Chin-Yi Cheng, Marco
  Fumero, and Kamal~Rahimi Malekshan.
\newblock Clip-forge: Towards zero-shot text-to-shape generation.
\newblock In {\em Proceedings of the IEEE/CVF Conference on Computer Vision and
  Pattern Recognition}, pages 18603--18613, 2022.

\bibitem{arda}
Arda Senocak, Hyeonggon Ryu, Junsik Kim, and In~So Kweon.
\newblock Less can be more: Sound source localization with a classification
  model.
\newblock In {\em WACV}, pages 3308--3317, 2022.

\bibitem{very}
Karen Simonyan and Andrew Zisserman.
\newblock Very deep convolutional networks for large-scale image recognition.
\newblock {\em arXiv preprint arXiv:1409.1556}, 2014.

\bibitem{tian}
Yapeng Tian, Jing Shi, Bochen Li, Zhiyao Duan, and Chenliang Xu.
\newblock Audio-visual event localization in unconstrained videos.
\newblock In {\em ECCV}, pages 247--263, 2018.

\bibitem{tsi}
Antigoni Tsiami, Panagiotis~Paraskevas Filntisis, Niki Efthymiou, Petros
  Koutras, Gerasimos Potamianos, and Petros Maragos.
\newblock Far-field audio-visual scene perception of multi-party human-robot
  interaction for children and adults.
\newblock In {\em ICAASP}, pages 6568--6572. IEEE, 2018.

\bibitem{clip2}
Can Wang, Menglei Chai, Mingming He, Dongdong Chen, and Jing Liao.
\newblock Clip-nerf: Text-and-image driven manipulation of neural radiance
  fields.
\newblock In {\em Proceedings of the IEEE/CVF Conference on Computer Vision and
  Pattern Recognition}, pages 3835--3844, 2022.

\bibitem{wu2}
Yifan Wu, Roshan Ayyalasomayajula, Michael~J Bianco, Dinesh Bharadia, and Peter
  Gerstoft.
\newblock Sslide: Sound source localization for indoors based on deep learning.
\newblock In {\em ICASSP}, pages 4680--4684. IEEE, 2021.

\bibitem{wu}
Yu Wu, Linchao Zhu, Yan Yan, and Yi Yang.
\newblock Dual attention matching for audio-visual event localization.
\newblock In {\em ICCV}, pages 6292--6300, 2019.

\bibitem{xia}
Fei Xia, Amir~R Zamir, Zhiyang He, Alexander Sax, Jitendra Malik, and Silvio
  Savarese.
\newblock Gibson env: Real-world perception for embodied agents.
\newblock In {\em CVPR}, pages 9068--9079, 2018.

\bibitem{xu}
Haoming Xu, Runhao Zeng, Qingyao Wu, Mingkui Tan, and Chuang Gan.
\newblock Cross-modal relation-aware networks for audio-visual event
  localization.
\newblock In {\em Proceedings of the 28th ACM International Conference on
  Multimedia}, pages 3893--3901, 2020.

\bibitem{xu2}
Xudong Xu, Bo Dai, and Dahua Lin.
\newblock Recursive visual sound separation using minus-plus net.
\newblock In {\em ICCV}, pages 882--891, 2019.

\bibitem{xuan}
Hanyu Xuan, Lei Luo, Zhenyu Zhang, Jian Yang, and Yan Yan.
\newblock Discriminative cross-modality attention network for temporal
  inconsistent audio-visual event localization.
\newblock {\em IEEE Transactions on Image Processing}, 30:7878--7888, 2021.

\bibitem{xuan2}
Hanyu Xuan, Zhenyu Zhang, Shuo Chen, Jian Yang, and Yan Yan.
\newblock Cross-modal attention network for temporal inconsistent audio-visual
  event localization.
\newblock In {\em AAAI}, volume~34, pages 279--286, 2020.

\bibitem{clip3}
Renrui Zhang, Ziyu Guo, Wei Zhang, Kunchang Li, Xupeng Miao, Bin Cui, Yu Qiao,
  Peng Gao, and Hongsheng Li.
\newblock Pointclip: Point cloud understanding by clip.
\newblock In {\em Proceedings of the IEEE/CVF Conference on Computer Vision and
  Pattern Recognition}, pages 8552--8562, 2022.

\bibitem{zhao}
Hang Zhao, Chuang Gan, Andrew Rouditchenko, Carl Vondrick, Josh McDermott, and
  Antonio Torralba.
\newblock The sound of pixels.
\newblock In {\em ECCV}, pages 570--586, 2018.

\bibitem{zhou2}
Hang Zhou, Xudong Xu, Dahua Lin, Xiaogang Wang, and Ziwei Liu.
\newblock Sep-stereo: Visually guided stereophonic audio generation by
  associating source separation.
\newblock In {\em ECCV}, pages 52--69. Springer, 2020.

\bibitem{zhou}
Jinxing Zhou, Liang Zheng, Yiran Zhong, Shijie Hao, and Meng Wang.
\newblock Positive sample propagation along the audio-visual event line.
\newblock In {\em CVPR}, pages 8436--8444, 2021.

\bibitem{zhu}
Zheng Zhu, Wei Wu, Wei Zou, and Junjie Yan.
\newblock End-to-end flow correlation tracking with spatial-temporal attention.
\newblock In {\em CVPR}, pages 548--557, 2018.

\end{thebibliography}
}

\end{document}